# A novel dataset and a two-stage mitosis nuclei detection method based on hybrid anchor branch

Huadeng Wang[1, 2], Hao Xu[1], Bingbing Li[3, *], Xipeng Pan[1, 2, *], Lingqi Zeng[1], Rushi Lan[1, 2], Xiaonan Luo[1, 2]

[1]School of Computer Science and Information Security, Guilin University of Electronic Technology, Guilin, 541004, China

[2]Guangxi Key Laboratory of Image and Graphic Intelligent Processing, Guilin, 541004, China

[3]Department of Pathology, Ganzhou Municipal Hospital, Ganzhou, 341000, China

E-mail: whd@guet.edu.cn, pxp201@guet.edu.cn

**Abstract**

Mitosis detection is one of the challenging problems in computational pathology, and mitotic count is an important index of cancer grading for pathologists. However, current counts of mitotic nuclei rely on pathologists looking microscopically at the number of mitotic nuclei in hot spots, which is subjective and time-consuming. In this paper, we propose a two-stage cascaded network, named FoCasNet, for mitosis detection. In the first stage, a detection network named $M_{det}$ is proposed to detect as many mitoses as possible. In the second stage, a classification network $M_{class}$ is proposed to refine the results of the first stage. In addition, the attention mechanism, normalization method, and hybrid anchor branch classification subnet are introduced to improve the overall detection performance. Our method achieves the current highest F1-score of 0.888 on the public dataset ICPR 2012. We also evaluated our method on the GZMH dataset released by our research team for the first time and reached the highest F1-score of 0.563, which is also better than multiple classic detection networks widely used at present. It confirmed the effectiveness and generalization of our method. The code will be available at: https://github.com/antifen/mitosis-nuclei-detection.

**Key words**：Pathological image; Mitosis nuclei detection; Cascade network; Hybrid anchor branch; Group normalization

## 1 Introduction

Breast cancer has become the most common cancer in the world for the first time (11.7% of new cases), with the highest morbidity of 24.2% and mortality of 15% among women [1]. In the current diagnosis of breast cancer, the Nottingham grading system is the most recommended breast cancer grading system by the World Health Organization (WHO) [2,3]. This system evaluates three morphological features: the pleomorphism of the nucleus, mitosis count, and gland formation on histopathological slides. Among them, the mitosis count is the most important one, because the proliferation of cancer is mainly controlled by cell division. Currently, hematoxylin and eosin (H&E) are used to stain histopathological slides in clinical research, and H&E-stained histopathological images can visually show components of cells and tissue structures. In automatic mitosis detection based on deep learning, pathologists need to manually label the observed mitosis nuclei on a high-power field (HPF), which is a subjective and time-consuming task and requires extensive experience and professional equipment. Therefore, developing an automatic detection method for mitosis nuclei will save time and labor resources for pathological diagnosis, and will also increase the reliability of the pathological analysis.

However, automated detection of mitosis nuclei is a challenging task for several reasons. First, the complexity of mitosis. Mitosis is divided into four phases (prophase, metaphase, anaphase, and telophase), and the nuclei shape and texture of each phase are very different, as shown in Fig. 1. In addition, there may be many other cells such as apoptotic cells and lymphocytes that have the same appearance as mitosis nuclei in H&E-stained images. We call them hard samples, which are very easy to be falsely detected as mitosis nuclei. Second, the imbalance of positive and negative samples. Compared with other non-mitosis nuclei, the number of mitosis nuclei in a single HPF is very small, which makes it difficult to extract effective features due to class imbalance. Third, datasets are insufficient, but most automatic detection methods rely on large amounts of data to support the accuracy of model training. Most of the current public datasets come from medical image processing research challenges, and the image quality is higher than the data directly from the hospital, but the total amount is smaller. Moreover, the structure and morphology of mitosis nuclei are variable, and the datasets cannot cover all pathological types.

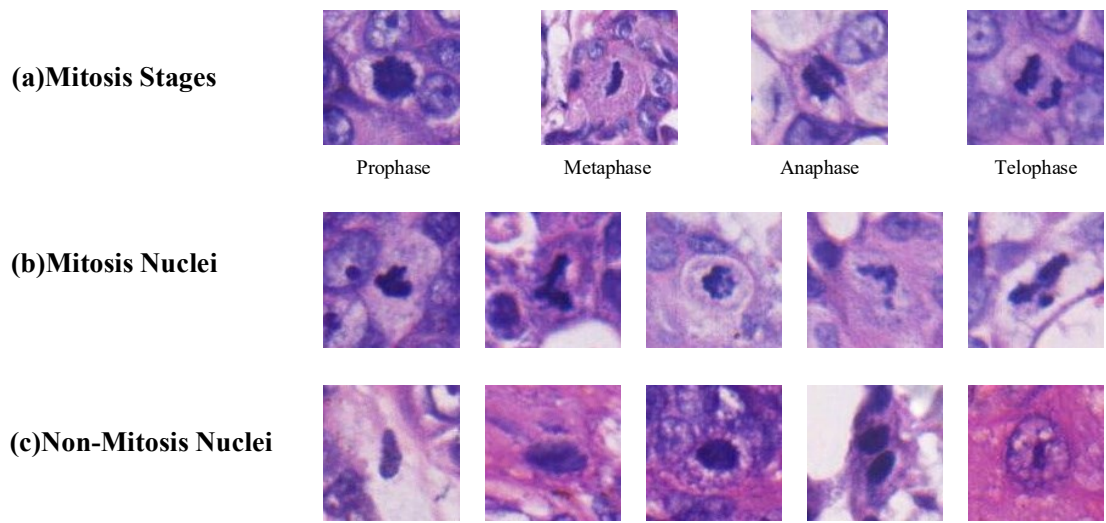

Figure 1. (a) Diagram of the four stages of mitosis nuclei; (b) mitosis nuclei (c)non-mitotic nuclei

Given the above shortcomings, the introduction of computer-assisted automatic detection, especially the deep learning method, has attracted more and more attention of researchers in recent years, which can help reduce the workload of the pathologist and improve the efficiency of diagnosis. At present, several international challenges have been held to study the specific application of deep learning methods in the mitosis detection of breast cancer. For example, ICPR Mitosis Detection Challenge in 2012 [4], AMIDA13 of MICCAI Grand challenge in 2013 [5], ICPR MITOS-ATYPIA Challenge in 2014 [6], and TUPAC16 Challenge of MICCAI in 2016 [7]. These challenges attracted many researchers to participate in, and many excellent methods based on these datasets appeared. For example, C. Li et al. designed the Deep mitosis network [8], Sebai et al. proposed PartMitosis [9], and so on. However, the datasets of these challenges are selected by the organizers and the data providers, and there are still some differences with the data used in clinical applications in hospitals. In other words, most current methods rarely perform very well in clinical data.

Aiming at the above issues, we propose the FoCasNet model, which is an improved two-stage network that can detect mitosis nuclei more accurately. Current detection or segmentation networks have limited performance, and it is difficult for single-stage networks to accurately identify mitosis nuclei. Therefore, we first use the proposed rough detection network $M_{det}$, which sets a low threshold to detect all mitosis nuclei as much as possible. Next, we use the more sensitive classification network $M_{class}$ to filter a large number of FP samples and the hard samples generated by the first stage. We evaluate our method on the public dataset ICPR 2012 and a novel dataset, GZMH Dataset, first published in collaboration with the researcher from Ganzhou Municipal Hospital, China. In terms of detection accuracy, our method achieves the highest F1-score of 0.888 on the 2012 ICPR MITOSIS Dataset. At the same time, we also achieve the highest F1-score of 0.563 on the GZMH Dataset.

The main contributions of this paper are as follows.

1) We propose an end-to-end cascade network for the automatic detection of mitosis nuclei in breast cancer, and our method achieves the best performance compared with other methods on both public dataset and our first published hospital clinical dataset.

2) An improved residual network backbone is proposed to re-encode features to enhance the ability of the model to detect mitosis nuclei.

3) A hybrid anchor branch method is proposed to make the network adaptively select the optimal scale, which improves the detection accuracy of the target object.

## 2 Related Work

In the 1980s, some researchers conducted a feasibility study on image processing of mitosis in breast cancer slices [10]. Then at the end of the last century, T. Kate et al. [11] proposed a mitosis calculation method under Feulgen staining. With the development of the times, the current detection methods of mitotic nuclei can be divided into traditional methods and deep learning methods. The former relies more on professional image processing technology, and generally requires researchers to manually distinguish and extract target features, and may be more complicated. With the support of rapidly developing computer software and hardware, the latter often takes less time with higher precision, and has become the primary choice. At present, more and more researchers use deep learning technology to automatically extract target features and have achieved impressive achievements.

### 2.1 The traditional method of mitosis nuclei detection

In traditional methods, image processing techniques are required to automate the detection by manually designing and selecting features. The morphological features such as area, perimeter, eccentricity, long axis length, short axis length, equivalent diameter, etc., and the statistical features such as mean, median, the variance of each color channel, color histogram features, color scale, etc., are selected to train classifiers to distinguish mitotic and non-mitotic nuclei on histopathological slides. Manually extracting image features is the first attempt to automatically detect mitosis nuclei in histopathological slides [12, 13, 14]. C. Sommer et al. [12] first used a pixel-level classifier to cut out candidate mitotic cells, and then used the texture, intensity, shape and other features of the training samples to train SVM and classify mitotic cells. H. Irshad et al. [13] used Gaussian Laplacian, threshold processing, etc. on the image to detect and segment candidate objects in the candidate detection stage. In the candidate classification stage, a total of 143 morphological features were extracted and finally classified using a decision tree classifier. FB. Tek et al. [14] investigated a set of general features (including granularity, color, and channel), and then used it to train a cascaded AdaBoosts classifier. The result showed that the features of granular structure and color change can be used for mitosis detection. However, due to the variety of shapes and textures of mitosis nuclei, it is difficult to customize hand-crafted feature-based methods for mitosis detection tasks, and the results of the above methods are not satisfactory.

**2.2 Deep learning-based mitosis nuclei detection methods**

Since 2010, the ImageNet Large-Scale Visual Recognition Challenge (ILSVRC) has been held every year, and many excellent algorithms have appeared, which has led to a huge development in the field of computer vision [15]. The introduction of CNN has greatly improved the effect of computer vision tasks and is widely used in the fields of classification, segmentation, and detection. Meanwhile, many new algorithms applied in the field of medical image processing have been shown to outperform the results of state-of-the-art hand-crafted feature-based classification methods [16].

Many algorithms based on deep learning have been successful in the task of detecting mitosis from histopathology images [17,18,19]. E. Zerhouni et al. [17] proposed a workflow based on Wide Residual Network, the model was trained to classify each pixel on an image using a pixel-centric patch as context, and then post-process the network output to filter out the noise and select the true mitosis. This method finally ranked second in the MICCAI TUPAC 2016 mitosis detection competition. N. Wahab et al. [18] first used the concept of transfer learning, and then used a pre-trained convolutional neural network (CNN) for segmentation. They secondly used another hybrid CNN (with weight transfer and custom layers) for mitotic classification. Fine-tuning based transfer learning reduced training time, provided good initial weights, and improved the F-score. H. Chen et al. [19] proposed a cascaded framework of deep CNNs, which consists of two parts: a fully convolutional network for highlighting and retrieving mitotic candidates, and a deep CNN classifier for better distinguish between mitosis nuclei and cells with a similar appearance. The method achieves excellent performance on the 2014 ICPR MITOSIS dataset with an F-score of 0.442.

The introduction of region-based convolutional neural networks (RCNN) [20] has greatly improved the performance of object detection algorithms. The first step of RCNN is to extract features from object region proposals generated by selective search methods [21]. These features are then used to train a set of SVMs for object class prediction and a bounding box regressor for object location estimation. However, RCNN takes a long time to run because it uses the outer region proposal method to generate proposals, and its convolutional forward pass is performed individually for each proposal. To solve this problem, Faster-RCNN [22] uses a Region Proposal Network (RPN) for object proposal prediction, which shares the same convolutional layers as the detection network, and uses a Region of Interest (RoI) pooling layer from each object extracting features from proposals for generating feature maps for the entire input image. Later, object detection networks such as RetinaNet [23] using Focal Loss and Mask RCNN [24] using a multi-task branch to enhance classification performance also emerged. The researchers have applied these novel networks to the field of mitosis detection, or used cascaded networks, all of which significantly improved the detection performance.

C.Li et al.[8] designed the Deepmitosis network to detect mitosis nuclei, which consists of three parts: a deep segmentation network (Deepseg), a deep detection network (Deepdet), and a deep verification network (Deepver). The first network is a segmentation network, where FCN extracts features and generates corresponding bounding boxes. The detection stage uses an RPN network to generate proposals at each location, which are then classified using a region-based classifier. The verification network is based on ResNet and reinforced with hard-negative examples. H. Lei et al. [25] used a two-stage approach to detect mitosis. They use VGG-16 as the backbone network to obtain deep and shallow features of the image and fuse them. The fused features are upsampled and fed into the next modified R-CNN network. Two classification branches are designed to improve detection accuracy. One is to generate fixed-size features and give prediction scores through ROI pooling layers. The other is to use a position-sensitive ROI pooling layer to generate maps corresponding to different positions and obtain scores using a voting mechanism, comprehensively considering both scores as the final output. Sebai et al. [26] proposed a multi-task deep learning framework MaskMitosis, which is mainly used for object detection and instance segmentation based on Mask RCNN. It is used as a detection network to perform mitotic localization and classification in fully annotated mitosis datasets (i.e. pixel-level annotated dataset), and it is used as a segmentation network to estimate masks for weakly annotated mitosis datasets (with only centroid pixel labels). Finally, it achieves the highest F-score of 0.863 on the 2012 ICPR dataset, while achieving an F-score of 0.475 on the 2014 ICPR dataset, outperforming all state-of-the-art mitosis detection methods.

**3 Method**

The overall structure of our FoCasNet model proposed in this paper is shown in Fig. 2, which mainly includes two convolutional neural networks. The former is an object detection network, and the latter is a classification network, they are combined in a cascade manner. The object detection network can retrieve as many mitosis nuclei as possible while maintaining robustness. We call it the rough detection network $M_{det}$ which outputs the anchor box coordinates and the corresponding probability values of all considered positive samples, that is mitosis nuclei candidates. The results of the detection network are post-processed into 64x64 image patches, and then sent to the cascaded classification network, which we call the elaborate classification network $M_{class}$. The classification network will learn the data distribution characteristics through training and can obtain very accurate classification results. Finally, FoCasNet combines the results of a rough detection network and an elaborate classification network to obtain complete image-level detection and classification results. In addition, the input of the classification network is the result of the detection network, which is consistent with the training and can ensure that the classification network learns the data distribution characteristics of the dataset and successfully uses them. In the feature detection stage, we use CBAM (Convolutional Block Attention Module) to re-encode spatial and channel features to improve the weight of key features, use group normalization and weight standardization to normalize weight and layer input output to avoid the constraints of batch size, speed up the convergence of the model, and use feature pyramid net to fuse multiscale features of convolution layers to better detect small objects, and finally, the rough detection network $M_{det}$ can detect the mitosis nuclei to the maximum extent, to facilitate the subsequent cascade classification network $M_{class}$ for classification and screening and improve the overall network performance.

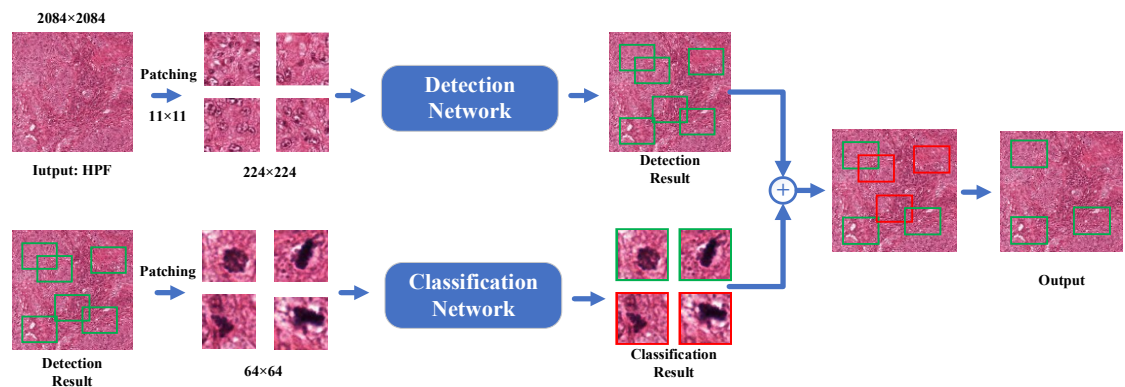

Figure 2. The overall architecture of the FoCasNet

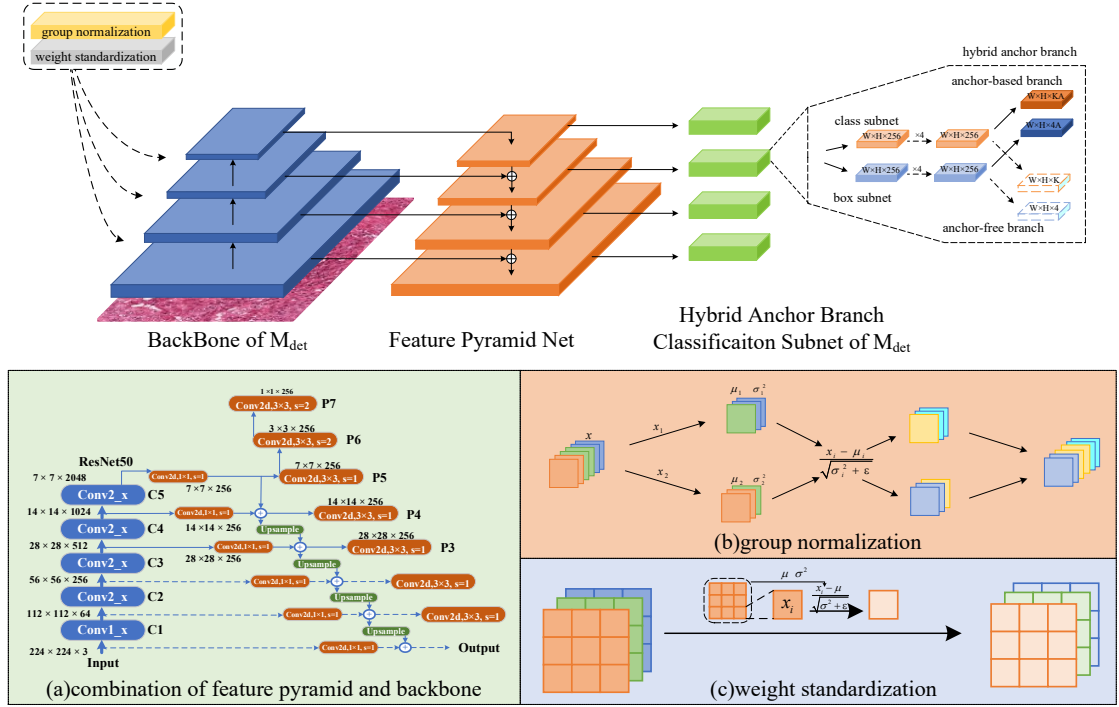

Figure 3. The rough detection network $M_{det}$

## 3.1 The rough detection network $M_{det}$

Our task in the first stage is to detect as many mitosis nuclei as possible, however, the detection task is difficult due to the uneven distribution of mitosis nuclei and the very small proportion of the whole. Therefore, we propose an improved object detection network called the rough detection network $M_{det}$, whose specific network structure is shown in Fig. 3. First, we use the improved backbone network to extract mitosis nuclei, which contains two parts: the feature extraction residual network and the feature pyramid network (Feature Pyramid Net), followed by a hybrid anchor branch subnet to detect and identify mitosis nuclei. Finally, the output is a set of bounding boxes $\{(x_1, y_1, x_2, y_2, W_1), \ldots, (x_{n1}, y_{n1}, x_{n2}, y_{n2}, W_n)\}$, where the elements in each tuple correspond to the coordinates of the top-left point $(x_1, y_1)$, the coordinates of the bottom-right point $(x_2, y_2)$, and the corresponding positive sample probability value W of the anchor where the prediction object is located, respectively. Next, we elaborate on each part of the $M_{det}$.

### A. The improved backbone network

We chose ResNet-50 as the backbone network for feature extraction, because ResNet-50 can effectively avoid the problem of gradient vanishing by using residual structure, speeding up the training process, and having deeper layers, which can better extract the target features. However, due to the small target and small proportion of mitosis nuclei, ResNet-50 cannot be used directly to solve the problem, so we targeted to improve the backbone network ResNet-50.

**Convolutional Block Attention Module.** To enable the backbone network to better extract object features, we add CBAM between the convolutional layers. It is a simple but effective feedforward neural network attention module with almost negligible overhead. CBAM uses a hybrid attention mechanism to focus on both the spatial and channel dimensions of features and re-encodes and combines the features of these two dimensions during backpropagation, which, for this task, highlights certain ROI (Region of Interest) regions and some important features and enhance their weights to facilitate the detection performance of the backbone network. We assume that the output feature map of a certain layer of convolution is $M \in R^{C \times H \times W}$, and the attention weight matrix generated by CBAM is $M' \in R^{C \times 1 \times 1}$. The process is shown in equations (1) and (2):

$$M'_c = \sigma(C_1(C_2(M_{avg}^c)) + C_1(C_2(M_{avg}^c))) \qquad (1)$$

$$M'_s = \sigma(f^{7 \times 7}([M_{avg}^c; M_{avg}^c])) \qquad (2)$$

where $M'_c$ is the attention weight matrix of the feature dimension, $M'_s$ is the attention weight matrix of the spatial dimension, $M_{avg}^c$ denotes the features of the M after average pooling, $M_{max}^c$ denotes the features of M after max pooling, $C_1$ and $C_2$ denotes the weights corresponding to the two fully connected layers, $f^{7 \times 7}$ denotes the convolution operation (the kernel size is 7×7), and $\sigma$ denotes the activation function.

Next, multiply the obtained attention weight matrix M' by the initial input feature M to obtain the final output feature map M", which is expressed as Eq. (3).

$$M'' = M' \otimes M \qquad (3)$$

where $\otimes$ represents the element-wise multiplication of the matrix.

```
def GroupNorm(x, gamma, beta, G, eps=1e-5):
    # x: input features with shape [N, C, H, W]
    # gamma, beta: scale and offset, with shape [1, C, 1, 1]
    # G: number of groups
    # mean, var with shape [N, G, 1, 1, 1]

    N, C, H, W= x.shape
    x = torch.reshape(input=x,  shape=[N,G, C// G, H, W])

    mean = torch.mean(input=x, dim=[2,3,4], keepdim=True)
    var = torch.var(input=x, dim=[2,3,4], keepdim=True)

    x = (x - mean) / torch.sqrt(var + eps)
    x = torch.reshape(input=x, shape=[ N, C, H, W])
    return x *gamma + beta
```

Figure 4. Python code of Group Normalization based on PyTorch

**Group Normalization and Weight Standardization.** In order to utilize the feature information of the dataset more effectively, we also use GN (Group Normalization) and WS (Weight Standardization) techniques. Without normalization, researchers often need to carefully specify the learning rate and its decay strategy, and it is also easy to cause gradient vanishing or gradient explosion problems, which makes the overall training process slow and difficult to converge. At present, the most widely used normalization technology in the image processing field is BN (Batch Normalization), however, its reliance on a large batch size makes it not suitable for all cases. In this task, we use the oversampling technique to preprocess the training set, and due to the limitation of the GPU memory, it is impossible to use a larger batch size. Therefore, we use Group Normalization instead of the BN layer, both of which are normalization techniques for activation, and experiments show that the fusion of GN and WS achieves better results than using BN alone [27].

The GN layer lies between the convolutional layer and the nonlinear activation layer.

Fig. 4 shows the Pytorch-based code. We assume that the feature map output from the previous layer of convolution is a 4D vector [N, C, H, W], where N represents the batch size, C represents the number of feature channels, H represents height, and W represents width. Then, GN groups the feature channels for a certain instance of the feature map, and then normalizes the channels, heights, and widths within the group, and continues for the whole batch. In this way, it not only avoids relying on large batch size, but also ensures that the normalization effect is close to the traditional normalization method.

Unlike GN, weight standardization is in another dimension, that is, normalization for weights. Since the inputs and layer outputs have been normalized, there are also weights left, which sometimes vary significantly due to outlier inputs, affecting the overall training process. By normalizing the weights, i.e., for the convolutional layers, a smoother loss and more stable training can be achieved. At the same time, directly adjusting the weight can also avoid the dependence on batch size. WS focuses on normalizing each kernel of the convolution layer. The output of the layer is n channels, n convolution kernels are required to correspond to it, and n normalization is also required. Similar to GN, the Pytorch-based code is shown in Fig. 5.

```
def WeightStand(w, eps=1e-5):
    # w: input features shape [Cin, Cout, kernel_size, kernel_size]
    # mean, var with shape [1, Cout, 1, 1]

    mean = torch.mean(input=w, dim=[0,2,3], keepdim=True)
    var = torch.var(input=w, dim=[0,2,3], keepdim=True)

    w = (w - mean) / torch.sqrt(var + eps)
    return w
```

Figure 5. Python code of Weight Standardization based on PyTorch

**Feature Pyramid Net.** Given the small size of mitosis nuclei and the difficulty in detecting them, we introduce a feature pyramid network (Feature Pyramid Net) to fuse the multi-layer features of the backbone network and improve the feature extraction capability. The feature extraction stage of object detection used to only use the last feature map of the last stage. Although the feature map generated by deep layer convolution assembles more high-level features, it is not suitable for this task. Mitosis nuclei is a small object whose feature information is mainly retained in the shallow layer, often losing a lot of semantic information as the convolution layer deepens, and because its pixel proportion is too small, its proportion in the output feature map of the deep convolution layer is even smaller, which causes the object detection network to fail to detect mitosis nuclei well. Therefore, we add a feature pyramid layer behind the backbone network, whose specific structure is shown in Fig. 3. We use a $1 \times 1$ convolution kernel after each convolution layer of the backbone network to change the number of channels, and then pass the high-level features to the low-level features through upsampling and fuse them, which results in high-resolution, strongly semantic features that facilitate the detection of small objects, namely mitosis nuclei.

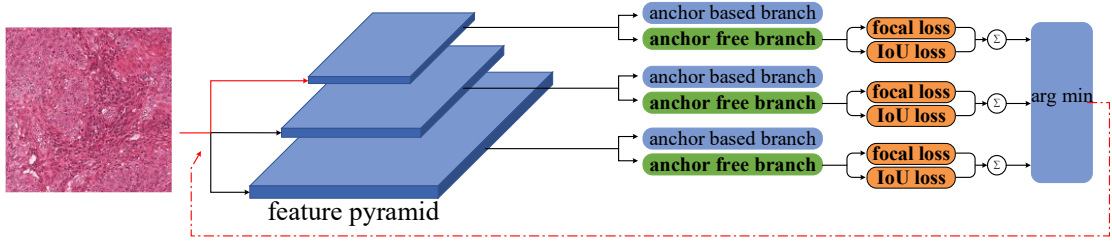

Figure 6. Feature Selection Process

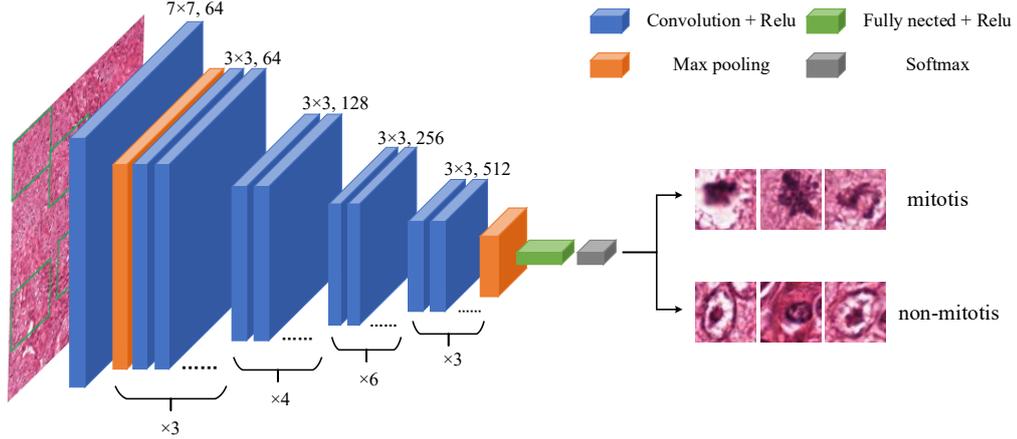

Figure 7. The elaborate classification network $M_{class}$

### B. Classification subnet based on hybrid anchor branch

After the backbone network extracts the object features, we directly use the classification subnet based on the hybrid anchor branch to achieve detection. The difference is that the anchor-based branch consists of 3 convolutional layers with shared weights, which predict the class to which each anchor belongs and the object bounding box regression parameters respectively. The anchor-free branch has another purpose. It is used to solve the scale dimension problem caused by the use of FPN in the backbone network. Next, we will elaborate on each part of the classification subnet.

**Anchor-based branch.** The specific structure of this branch is shown in Fig. 3, which mainly contains two subnets: Classification Subnet and Box Regression Subnet. We assume that the detection task has a total of C categories and feature maps corresponding to A anchor boxes for each spatial position. The Classification Subnet is a small FCN network attached to the FPN, and the weights are shared with the FPN, with the main purpose of classifying the anchor boxes. It uses four 3x3 convolutional layers, each containing C convolution kernels, corresponding to the feature map with C channels generated by FPN, and then uses the Relu activation function to increase the non-linearity, followed by the same 3×3 convolutional layer to filter KA anchor boxes, and finally uses sigmoid activation to output KA binary predictions for each spatial location of the feature map. Here, K refers to the number of categories of object detection, excluding the background. In this task, C = 2, A = 9, and K = 1. Similarly, the Box Regression Subnet is also a small FCN network attached to the FPN. It is performed in parallel with the Classification Subnet during propagation. The main task is to regress the bounding box parameters to be closer to the ground truth. The network settings of both are mostly the same. The difference is that the last layer of convolutional outputs 4A parameters, which correspond to the 4 offsets of the A anchor box at each spatial position of the feature map from the ground truth box coordinates. Therefore, the Anchor-based branch predicts the class and location information of each anchor box based on these two branches, and finally outputs them.

**Anchor-free branch.** The rough detection network $M_{det}$ can achieve the whole detection process with an Anchor-based branch alone, but it may cause scale dimension problems due to the use of FPN. In the traditional object detection network, the feature map of a certain size should predict which size range of objects is manually set. As a result, large objects will be predicted by the high-level feature map, while small objects will be predicted by the low-level feature layer. In some cases, this heuristic setting may not be optimal. For example, in this task, most of the mitosis nuclei are small objects. Based on the above description, the features of these samples are usually predicted by the shallow neural network, because the low-level feature layer contains more small object feature information, so it can be said that the prediction of mitosis nuclei by the shallow neural network is in line with the expectation and optimal. However, the mitosis nuclei of some patients may change to different degrees in clinical practice, which is mainly reflected in the increase or decrease of the overall size. In this case, it is not the most effective to use the original feature layer to predict the location of the mitotic image. Therefore, we introduced an Anchor-free branch to enable the input to select the feature layer adaptively through training.

The specific structure is shown in Fig. 3, which is very simple overall. Since it does not rely on the anchor box, for each spatial position of the feature map, the output size of class prediction and bounding box are $W \times H \times K$, and $W \times H \times 4$, respectively. After training, the process of instance adaptive selection of feature layers is shown in Fig. 6. Once each instance passes the feature pyramid FPN, the loss of the two branches of the Anchor-free branch is calculated at each layer, where the classification subnet uses focal loss, the box regression subnet uses IoU loss, and then the two losses are summed, and the feature layer with the lowest loss is finally used to predict the instances. Since this branch is trainable, the optimal feature layer can be automatically selected for each input instance as the model converges.

### 3.2 The elaborate classification network $M_{class}$

This network aims at screening out too many false mitosis nuclei. Since the rough detection network $M_{det}$ in the first stage focuses on detecting as many mitosis nuclei as possible, that is, increasing the Recall value, this will increase the number of false positive FP at the same time. Therefore, this stage uses the elaborate classification network $M_{class}$ with higher feature extraction ability to screen the mitotic candidates generated by $M_{det}$, especially focusing on identifying false positive samples, and finally combines the output of the previous stage to obtain the final recognition result. Its network structure is shown in Fig. 7, and we use the residual network ResNet-34 as the overall classification network.

Although ResNet has been shown to achieve excellent results in many fields, it is not very effective in the medical field, especially in the classification of mitosis nuclei. This is partly since the current dataset is too small and the size of the mitosis nuclei is small, which makes it difficult to extract effective features. Another part is that there are many samples similar to the mitosis nuclei in the dataset (we call it hard example), which hinders the identification of the true mitosis nuclei. To this end, we have made the following improvements.

(a) Use data augmentation. Data augmentation is a common approach when training small datasets, where we augment the dataset with spatial transformations (rotation, translation, flipping, cropping, etc.) and color transformations (contrast, saturation, Gaussian noise, etc.).

(b) Use pre-trained weights. Studies have shown that ResNet classifiers trained on the ImageNet large-scale natural image dataset can facilitate related problems in other fields. Although natural images and medical images for this task have large differences in high-level features, they are similar in low-level details. Therefore, we use pre-training weights, which is equivalent to pre-training the classification network, which optimizes the feature extraction ability on the one hand and speeds up the training speed on the other.

(c) Balance of positive and negative samples. In this task, positive sample mitosis nuclei account for a small proportion of the dataset, while the negative sample composed of background and hard example is relatively large. This leads to an imbalance of positive and negative samples. Therefore, we use the oversampling technique to massively expand the number of positive samples. The specific method is to randomly select any pixel value of a positive sample as a patch centroid for cropping, while the negative samples are randomly cropped in the background.

(d) Hard Example Mining. The correct identification of hard examples is also one of the difficulties in the classification stage because they are similar to the true mitosis nuclei in appearance. We found that although the results of the first stage detector $M_{det}$ contained many FPs, most of these FPs were nuclei with similar morphology to the mitosis nuclei, which was highly coincident with hard examples. Therefore, we put the FP part of the detection results of $M_{det}$ into the negative samples of the classifier $M_{class}$ as prior knowledge. In this way, the classification network can better learn the feature distribution of hard examples, and it also ensures the logical unity of the training and testing processes.

(e) More careful parameter tuning. We found that following the initial parameter settings of ResNet does not fit this task well. After a lot of experiments and comparisons, the final patch size with an input of 128 and a patch size of 224 in the test can achieve better results.

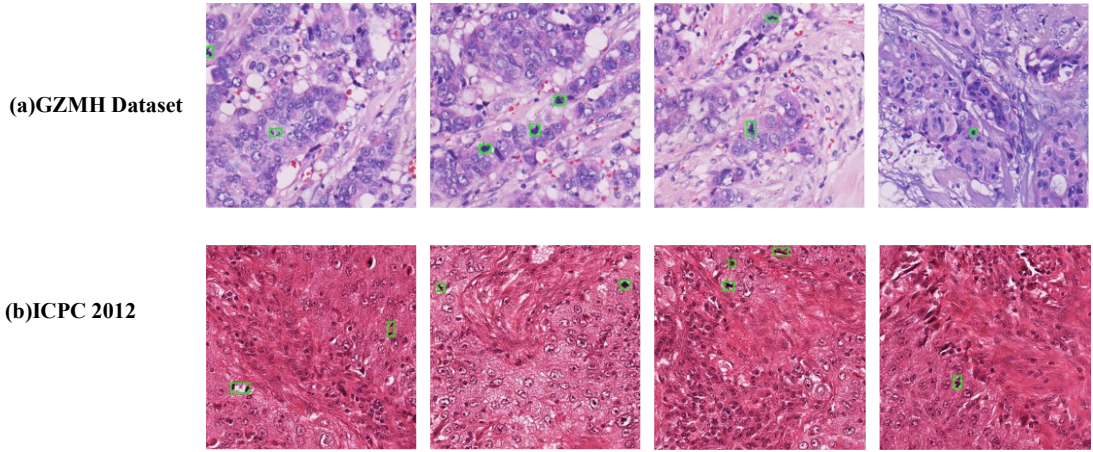

Figure 8. Detection results of different datasets highlighted in green

## 4 Experiment
### 4.1 Dataset and Preprocessing

In this paper, the performance of the proposed model is tested on a publicly available mitosis dataset, ICPR 2012 MITOSIS. The dataset consists of 50 HPFs acquired with an AperioXT scanner at ×40 magnification at a spatial resolution of 0.2456 μm /pixel. The size of each HPF is 2084x2084, corresponding to 512×512 μm$^2$ pixel values. More than 300 mitosis nuclei were annotated, and the pathologist precisely labeled every pixel of each mitosis nuclei (fully labeled dataset). Following the standard of the ICPR 2012 challenge, we use 35 HPFs for training and the remaining 15 for testing. Due to the large size of the HPF, direct input to the network is not conducive to the feature extraction of the neural network. We crop patches of 224×224 size in the first stage of detection to ensure that certain mitosis nuclei are in each patch. In the classification network, patches of 64×64 size are constructed, and the same is true for negative samples. The reason for the smaller patch size in the classification network is that we need finer object details and a larger proportion of pixels to correctly classify mitosis nuclei and other negative samples, including background and hard examples.

In addition, we also release and conduct validation experiments on the GZMH dataset, a dataset originating from Ganzhou Municipal Hospital in Jiangxi Province, China for the first time[1]. The dataset was provided by doctors in Ganzhou Municipal Hospital, which contains 55 WSIs from 22 different patients, scanned from a digital slice scanner (KF-PRO120) with a scan ratio of 40× and a resolution of 0.25um/pixel. Annotation is the bounding box coordinates used by the detection network, rather than pixel-level annotations. The annotation of this dataset was first completed by 3 pathologists with more than 5 years of working experience and then reviewed by 2 senior pathologists to ensure the correctness of the annotation. Also, we divide the dataset in such a way that the training set and test set are from different patients to avoid crossover. We used about 70% of them for training, 48 WSIs from a total of 20 patients, and the rest for testing, a total of 7 WSIs from 2 patients. In terms of data processing, we first manually select hot spots (regions with more mitosis nuclei), and then to ensure the consistency of preprocessing of different datasets, these regions are divided into HPFs of 2084×2084 size, and the subsequent operations are the same as the previous ones.

---

[1] The dataset is available at https://bit.ly/GZMH_Dataset.

The experimental environment is two Intel X4210R, 2.40GHz, 10-core, 20-thread CPUs with 256 GB of memory and two NVIDIA RTX 3090 GPUs, and the experiments are done on Ubuntu 16.04 operating system. The results of the two datasets mentioned above are shown in Fig. 8.

**4.2 Evaluation Metrics**

In this task, we focus on the number of mitosis nuclei in breast cancer for use in the downstream breast cancer grading task. Therefore, our evaluation metrics all rely on mitosis nuclei counts. According to the current consistency standard, we consider the model detection result as correct if the center coordinates of the anchor box are within 5um (20 pixels) of the centroid of the ground truth. We use TP (True Positive) to represent the number of mitosis nuclei correctly in all detection results, FP (False Positive) to represent the number of mitosis nuclei detected incorrectly in all detection results, and FN (False Negative) to represent the number of mitosis nuclei not detected by the network. We evaluate the performance of the model using three different detection metrics, including Precision, Recall, and F1-score, whose formulas are shown in (4)-(6):

$$Precision = \frac{TP}{TP+FP} \quad (4)$$

$$Recall = \frac{TP}{TP+FN} \quad (5)$$

$$F1 - score = \frac{2\times(Precision + Recall)}{Precision \times Recall} \quad (6)$$

**Table 1.** Ablation study on the ICPR 2012 dataset

| Method (All with classification network ResNet34) | Precision | Recall | F1-score |
|---|---|---|---|
| Backbone (ResNet50 + FPN) | 0.89 | 0.57 | 0.70 |
| Backbone + hybrid anchor branch | 0.92 | 0.64 | 0.75 |
| Backbone + hybrid anchor branch + CBAM | 0.87 | 0.72 | 0.79 |
| Backbone + hybrid anchor branch + GN & WS | 0.87 | 0.71 | 0.78 |
| Backbone + hybrid anchor branch + CBAM + GN & WS | 0.90 | 0.73 | 0.81 |
| Backbone + hybrid anchor branch + CBAM + GN & WS (Use $M_{class}$) | 0.86 | 0.92 | **0.89** |

**4.3 Ablation Experiment**

In this section, we validate the impact of different improvement methods on the performance of the FoCasNet model on the ICPR 2012 dataset. The results of these methods on Precision, Recall, and F1-score are shown in Table 1. It can be seen that various improvements have different effects on different parameters, but all have great improvements. Among them, the highest detection accuracy is achieved by using our hybrid anchor branch method, which is because each ground truth can be trained to find the most suitable feature layer for forward propagation, so the detection accuracy and recall are significantly improved. The subsequent improvements to the backbone network are CBAM and GN & WS, CBAM highlights important features, and GN & WS performs normalization operations, both of which can detect more mitosis nuclei under the premise of basically ensuring the accuracy, corresponding to the close recall values of the two. But unfortunately, the combination of these two operations failed to achieve a more obvious improvement. The above results are all generated by the second-stage classification network ResNet34, which is simply trained on the ImageNet dataset. Finally, we use an improved strategy, namely $M_{class}$, which significantly improves the accuracy of the classification network and reaches the best F1-score of 0.89.

**Table 2.** Performance comparison of different patch size

| Method | Precision | Recall | F1-score |
|---|---|---|---|
| 64×64 | 0.27 | 0.79 | 0.4 |
| 112×112 | 0.86 | 0.92 | **0.888** |
| 128×128 | 0.59 | 0.77 | 0.66 |
| 256×256 | 0.11 | 0.57 | 0.18 |

**4.4 Effects of different patch size in classification**

In this section, we verify the effect of the patch size generated by the post-processing of the first-stage $M_{det}$ on the overall model performance on the ICPR 2012 dataset. As mentioned in section 4.1, we use a patch size of 64×64 in the second stage for training classification network $M_{class}$. It is generally believed that using the same size images for both training and testing procedures tends to yield optimal results. However, in this task, we have other findings and the results are shown in Table 2. The overall processing flow of FoCasNet in the test time is to crop the detection results of the first stage into small square patches with the center of each anchor box for the second stage classification. Then we observed that using the default patch size of 64×64 is not as good as using 128×128, regardless of Precision, Recall or F1-score, but further expanding the size is counterproductive. This is due to the proportion of mitosis nuclei pixels in the patch is too large, and it is difficult for the classification network to judge the image-level classification results. Finally, we fluctuate around the value of 128 and find that the patch size of 112×112 is the most suitable for this task, which may imply that the best patch size for the classification network corresponding to the ICPR 2012 dataset is 112×112.

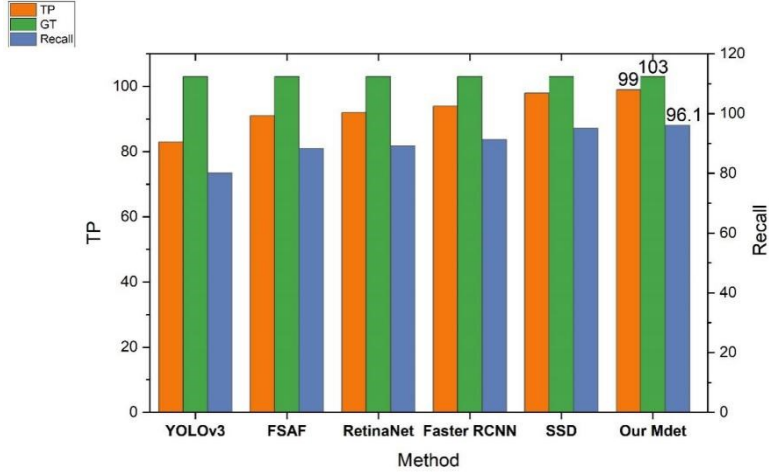

Figure 9. Performance comparison of different backbone networks (Detection Stage)

Table 5. Performance comparison of different backbone networks (Detection Stage)

| Method | TP | GT | Recall |
|---|---|---|---|
| YOLOv3 [28] | 83 | 103 | 0.801 |
| FSAF [29] | 91 | 103 | 0.883 |
| RetinaNet [23] | 92 | 103 | 0.893 |
| Faster RCNN [22] | 94 | 103 | 0.913 |
| SSD [30] | 98 | 103 | 0.951 |
| Our $M_{det}$ | 99 | 103 | **0.961** |

Table 6. Performance comparison of different backbone networks (Classification Stage)

| Method (Stage one) | Precision | Recall | F1-score |
|---|---|---|---|
| ResNet-18 | 0.81 | 0.73 | 0.770 |
| ResNet-34 | 0.86 | 0.92 | **0.888** |
| ResNet-50 | 0.86 | 0.75 | 0.800 |
| ResNet-101 | 0.80 | 0.70 | 0.746 |
| ResNet-152 | 0.82 | 0.70 | 0.754 |

**4.5 Selection of the backbone network**

In this section, we explore the impact of different backbone networks on mitosis detection and classification on the ICPR 2012 dataset. Since the backbone network also contributes to the detection performance, several experiments were conducted on FoCasNet using different backbone network models, i.e., YOLOv3 [28], FSAF [29], RetinaNet [23], Faster RCNN [22], and SSD [30]. Table 5 and Table 6 show the effect of the performance of different backbone networks on classification accuracy. Fig. 9 shows that our method detected the largest number of TPs (99) compared with other methods and therefore reached the highest Recall value of 0.961. Compared with some classical detection networks that are widely used, our improved rough detection network $M_{det}$ can identify the most positive samples, that is, the largest TP value, along with the corresponding largest Recall value. Furthermore, our theoretical arguments and experimental results show that ResNet34 is the best backbone model for the FoCasNet classifier. At first, we think of mitosis nuclei classification as a complex task, because accurately identifying mitosis nuclei and hard examples requires learning very fine-grained features, which often requires setting up multiple layers of convolutions to extract high-level features. However, during the experiment, we found that with the deepening of the network layers, various metrics were improved at the beginning, such as ResNet18 to ResNet34. But then the metrics decreased instead, and as the number of layers increased, they became lower and lower. We infer that this is because the mitosis nuclei are mainly distributed in the shallow layers, and the network with more layers lost these low-level features, which leads to the degradation of network performance. In other words, mitosis nuclei classification should be seen as a "simple" task.

**4.6 Performance comparison with other methods**

In this section, we compare the performance with state-of-the-art methods on the public dataset ICPR 2012. Three excellent methods appeared in the ICPR 2012 challenge, which are NEC [31], IPAL [32], and IDSIA [33]. Later, researchers proposed a method of fusing CNN and handcrafted features (HCF) [34]. The relative entropy maximizing scale space (REMSS) segmentation method and the RRF method based on the random forest (RF) classifier showed an F1-score of 0.823 [35]. CNN-based mitosis detection methods include CasCNN [19], DeepMitosis [8], and IDSIA. DeepMitosis proposed segmentation, detection, and validation models in 2018. Subsequently, the multi-task deep learning framework for object detection and instance segmentation Mask-RCNN [26] in 2020 achieves the current best F1-score for this dataset: 0,863. Fig. 10 shows that the Recall value of our proposed FoCasNet significantly exceeds all existing methods, and the F1-score also reaches the best level on the ICPR 2012 dataset, with an improvement of more than 2%, and the Precision also reaches a relatively high value.

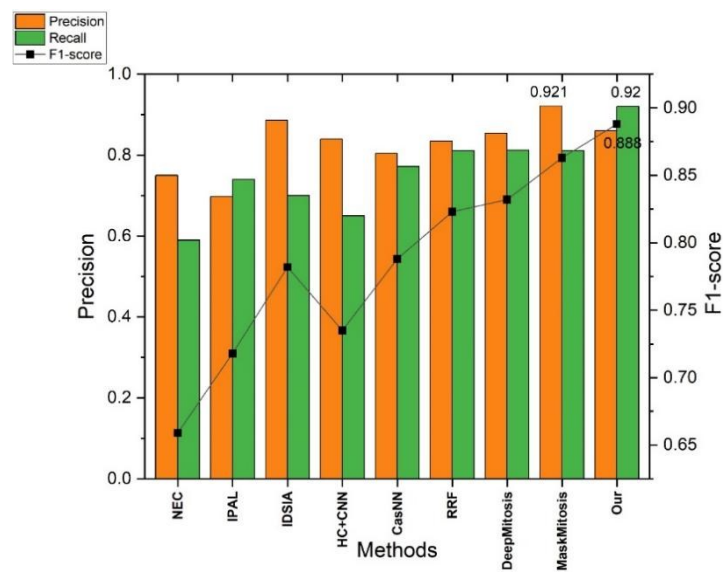

Figure 10. Performance comparison with other methods on the ICPR 2012 dataset

**Table 7.** Numerical performance comparison on the ICPR 2012 dataset

| Methods | Precision | Recall | F1-score |
|---|---|---|---|
| NEC [31] | 0.75 | 0.59 | 0.659 |
| IPAL [32] | 0.698 | 0.74 | 0.718 |
| IDSIA [33] | 0.886 | 0.70 | 0.782 |
| HC + CNN [34] | 0.84 | 0.65 | 0.735 |
| CasNN [19] | 0.804 | 0.772 | 0.788 |
| RRF [35] | 0.835 | 0.811 | 0.823 |
| DeepMitosis [8] | 0.854 | 0.812 | 0.832 |
| MaskMitosis [26] | **0.921** | 0.811 | 0.863 |
| Our FoCasNet | 0.86 | **0.92** | **0.888** |

**Table 8.** Performance Comparison With Other Methods On the GZMH Dataset

| Methods | Recall | F1-score |
|---|---|---|
| Faster R-CNN [22] | 0.497 | 0.173 |
| FSAF [29] | 0.348 | 0.436 |
| RetinaNet [23] | 0.491 | 0.476 |
| YOLOv3 [28] | 0.529 | 0.48 |
| SSD [30] | 0.489 | 0.511 |
| Our Method | **0.533** | **0.563** |

**4.7 Generalization Ability Verification**

In addition, we also released a novel GZMH dataset and verified the generalization ability of our model on this dataset. The advantage of this dataset is that it comes from clinical data, annotated by pathologists, and double-checked by senior pathologists to ensure the authenticity and accuracy of the data. This dataset is used by our research team for the first time. As shown in Table 8, we compare the performance with some reproducible classical detection networks. Among them, Faster R-CNN [22] is the leading two-stage object detection network, whose detection accuracy exceeds most networks on ImageNet, but does not perform well in this task. YOLOv3 [28] and RetinaNet [23] are single-stage object detection networks. The former is widely used in industry, and the focal loss proposed by the latter solves the problem of unbalanced positive and negative samples. Similarly, they achieve F1-score of 0.436 and 0.476 in this dataset, respectively, close to 50% detection accuracy. FSAF [29] is an improved version of RetinaNet and its performance is also similar. SSD [30] is also a single-stage network, which is characterized by a small model and fast training speed and is very widely used in industrial applications. It achieves an F1-score of 0.511, surpassing most other networks. This may confirm the conjecture of Section 4.5: the identification of mitosis nuclei may be more inclined to be a "simple" task because the performance of using simple networks often exceeds that of complex networks. Finally, our proposed FoCasNet achieves the highest F1- score of 0.563. It is worth noting that to verify the generalization ability of the method, we follow the processing flow of this method in the public dataset and use the same parameter setting as other classical networks.

**5 Conclusion**

This paper proposes a novel GZMH dataset and an improved two-stage mitosis nuclei detection method named FoCasNet. The first part of the FoCasNet is called the rough detection network $M_{det}$, whose task is to detect as many mitosis nuclei as possible, the second part is called the elaborate classification network $M_{class}$, and the task is to classify it finely screens the detection results of the first stage and remove false positives. We introduce methods such as attention mechanism, normalization technique, and feature pyramid to better extract object features, and use an innovative hybrid anchor branch classification subnet to achieve higher detection accuracy. In summary, while ensuring the detection accuracy, the detection Recall has been significantly improved, and finally achieved state-of-the-art results on the ICPR 2012 dataset. On GZMH, a clinical dataset from hospitals that we released for the first time, the performance of the proposed model is also superior to multiple reproducible classical detection networks. In the future, the detection capability of our proposed model can be further improved to enhance the performance and generalization ability of the model on clinical datasets.

**Reference**

[1] Latest global cancer data, Cancer burden rises to 19.3 million new cases and 10.0 million cancer deaths in 2020, https://www.iarc.fr/fr/news-events/latest-global-cancer-data-cancer-burden-rises-to-19-3-million-new-cases-and-10-0-million-cancer-deaths-in-2020/, 2020 (accessed22.01.10).

[2] C.W. Elston, I.O. Ellis. Pathological prognostic factors in breast cancer. I. The value of histological grade in breast cancer: experience from a large study with long-term follow-up, J. Histopathology. 19(5)(1991)403-410. https://doi.org/10.1111/j.1365-2559.1991.tb00229.x

[3] M. Veta, J.P.W. Pluim, P.J. Van Diest, et al. Breast cancer histopathology image analysis: A review, J. IEEE transactions on biomedical engineering. 61(5)(2014)1400-1411. https://doi.org/10.1109/TBME.2014.2303852

[4] R. Ludovic, R. Daniel, L. Nicolas, et al. Mitosis detection in breast cancer histological images An ICPR 2012 contest, J. Journal of pathology informatics. 4(1)(2013)8. https://doi.org/10.4103/2153-3539.112693

[5] M. Veta, P.J. Van Diest, S.M. Willems, et al. Assessment of algorithms for mitosis detection in breast cancer histopathology images, J. Medical image analysis. 20(1)(2015)237-248. https://doi.org/10.1016/j.media.2014.11.010

[6] MITOS-ATYPIA-14, Mitos-atypia-14-dataset, https://mitos-atypia-14.grand-challenge.org/dataset/, 2014 (accessed19.02.04).

[7] TUPAC16, Tumor-proliferation-assessment-challenge, http://tupac.tue-image.nl/, 2016(accessed 19.02.04).

[8] C. Li, X. Wang, W. Liu, et al. DeepMitosis: Mitosis detection via deep detection, verification and segmentation networks, J. Medical image analysis. 45(2018)121-133. https://doi.org/10.1016/j.media.2017.12.002

[9] M. Sebai, T. Wang, S.A. Al-Fadhli. PartMitosis: a partially supervised deep learning framework for mitosis detection in breast cancer histopathology images, J. IEEE Access. 8(2020)45133-45147. https://doi.org/10.1109/ACCESS.2020.2978754

[10] E.J. Kaman, A.W.M. Smeulders, P.W. Verbeek, et al. Image processing for mitoses in sections of breast cancer: a feasibility study, J. Cytometry: The Journal of the International Society for Analytical Cytology. 5(3)(1984)244-249. https://doi.org/10.1002/cyto.990050305

[11] T.K. Ten Kate, J.A.M. Belien, A.W.M. Smeulders, et al. Method for counting mitoses by image processing in Feulgen stained breast cancer sections, J. Cytometry: The Journal of the International Society for Analytical Cytology. 14(3)(1993): 241-250. https://doi.org/10.1002/cyto.990140302

[12] C. Sommer, L. Fiaschi, F.A. Hamprecht, et al. Learning-based mitotic cell detection in histopathological images, C. Proceedings of the 21st International Conference on Pattern Recognition (ICPR2012). IEEE(2012)2306-2309.

[13] H. Irshad. Automated mitosis detection in histopathology using morphological and multi-channel statistics features, J. Journal of pathology informatics. 4(1)(2013)10. https://doi.org/10.4103/2153-3539.112695

[14] F.B. Tek. Mitosis detection using generic features and an ensemble of cascade adaboosts, J. Journal of pathology informatics. 4(1)(2013)12. https://doi.org/10.4103/2153-3539.112697

[15] O. Russakovsky, J. Deng, H. Su, et al. Imagenet large scale visual recognition challenge, J. International journal of computer vision. 115(3)(2015)211-252. https://doi.org/10.1007/s11263-015-0816-y

[16] A. Janowczyk, A. Madabhushi. Deep learning for digital pathology image analysis: A comprehensive tutorial with selected use cases, J. Journal of pathology informatics. 7(1)(2016)29. https://doi.org/10.4103/2153-3539.186902

[17] E. Zerhouni, D. Lányi, M. Viana, et al. Wide residual networks for mitosis detection, C. 2017 IEEE 14th International Symposium on Biomedical Imaging (ISBI 2017). IEEE(2017)924-928. https://doi.org/10.1109/ISBI.2017.7950667

[18] N. Wahab, A. Khan, Y.S. Lee. Transfer learning based deep CNN for segmentation and detection of mitoses in breast cancer histopathological images, J. Microscopy. 68(3)(2019)216-233. https://doi.org/10.1093/jmicro/dfz002

[19] H. Chen, Q. Dou, X. Wang, et al. Mitosis detection in breast cancer histology images via deep cascaded networks, C. Thirtieth AAAI conference on artificial intelligence. 2016.

[20] R. Girshick, J. Donahue, T. Darrell, et al. Rich feature hierarchies for accurate object detection and semantic segmentation, C. Proceedings of the IEEE conference on computer vision and pattern recognition. (2014)580-587. https://doi.org/10.1109/CVPR.2014.81

[21] J.R.R. Uijlings, K.E.A. Van De Sande, T. Gevers, et al. Selective search for object recognition, J. International journal of computer vision. 104(2)(2013)154-171. https://doi.org/10.1007/s11263-013-0620-5

[22] S. Ren, K. He, R. Girshick, et al. Faster R-CNN: Towards real-time object detection

with region proposal networks, J. IEEE Transactions on Pattern Analysis & Machine Intelligence. 39(6)(2017)1137-1149. https://doi.org/10.1109/TPAMI.2016.2577031

[23] T.Y. Lin, P. Goyal, R. Girshick, et al. Focal loss for dense object detection, C. Proceedings of the IEEE international conference on computer vision. (2017) 2980-2988. https://doi.org/10.1109/TPAMI.2018.2858826

[24] K. He, G. Gkioxari, P. Dollár, et al. Mask r-cnn, C. Proceedings of the IEEE international conference on computer vision. (2017)2961-2969. https://doi.org/10.1109/ICCV.2017.322

[25] H. Lei, S. Liu, H. Xie, et al. An improved object detection method for mitosis detection, C. 2019 41st Annual International Conference of the IEEE Engineering in Medicine and Biology Society (EMBC). IEEE(2019)130-133. https://doi.org/10.1109/EMBC.2019.8857343

[26] M. Sebai, X. Wang, T. Wang. MaskMitosis: a deep learning framework for fully supervised, weakly supervised, and unsupervised mitosis detection in histopathology images, J. Medical & Biological Engineering & Computing. 58(7)(2020)1603-1623. https://doi.org/10.1007/s11517-020-02175-z

[27] S. Qiao, H. Wang, C. Liu, et al. Micro-batch training with batch-channel normalization and weight standardization, J. arXiv preprint. arXiv:1903.10520(2019). https://doi.org/10.48550/arXiv.1903.10520

[28] J. Redmon, A. Farhadi. Yolov3: An incremental improvement, J. arXiv preprint. arXiv:1804.02767(2018). https://doi.org/10.48550/arXiv.1804.02767

[29] C. Zhu, Y. He, M. Savvides. Feature selective anchor-free module for single-shot object detection, C. Proceedings of the IEEE/CVF conference on computer vision and pattern recognition. (2019)840-849. https://doi.org/10.1109/CVPR.2019.00093

[30] W. Liu, D. Anguelov, D. Erhan, et al. Ssd: Single shot multibox detector, C. European conference on computer vision. Springer, Cham. (2016)21-37. https://doi.org/10.1007/978-3-319-46448-0_2

[31] C.D. Malon, E. Cosatto. Classification of mitotic figures with convolutional neural networks and seeded blob features, J. Journal of pathology informatics. 4(1)(2013) 9. https://doi.org/10.4103/2153-3539.112694

[32] H. Irshad. Automated mitosis detection in histopathology using morphological and multi-channel statistics features, J. Journal of pathology informatics. 4(1)(2013)10. https://doi.org/10.4103/2153-3539.112695

[33] D.C. Cireşan, A. Giusti, L.M. Gambardella, et al. Mitosis detection in breast cancer histology images with deep neural networks, C. International conference on medical image computing and computer-assisted intervention. Springer, Berlin, Heidelberg. (2013)411-418. https://doi.org/10.1007/978-3-642-40763-5_51

[34] H. Wang, A. Cruz-Roa, A. Basavanhally, et al. Cascaded ensemble of convolutional neural networks and handcrafted features for mitosis detection, C. Medical Imaging 2014: Digital Pathology. SPIE, 9041(2014)66-75. https://doi.org/10.1117/12.2043902

[35] A. Paul, A. Dey, D.P. Mukherjee, et al. Regenerative random forest with automatic

feature selection to detect mitosis in histopathological breast cancer images, C. International Conference on Medical Image Computing and Computer-Assisted Intervention. Springer, Cham. (2015)94-102. https://doi.org/10.1007/978-3-319-24571-3_12